\PassOptionsToPackage{table}{xcolor}
\documentclass[runningheads]{llncs}
\usepackage[T1]{fontenc}
\usepackage{graphicx}
\usepackage{multirow, booktabs, array}
\usepackage{amssymb} 
\usepackage{amsmath} 
\usepackage[table]{xcolor}
\usepackage{subcaption}
\usepackage{enumitem}
\usepackage{mathtools}

\usepackage{amsthm}

\usepackage{algorithm}
\usepackage{algorithmic}

\definecolor{best}{RGB}{198,239,206}
\definecolor{second}{RGB}{189,215,238}
\definecolor{pos}{RGB}{0,128,0}
\definecolor{neg}{RGB}{178,34,34}
\definecolor{verylightblue}{RGB}{240, 248, 255}
\definecolor{softblue}{RGB}{220, 235, 250}

\begin{document}
\title{Bridging the Semantic Gap for Categorical Data Clustering via Large Language Models}
\titlerunning{Bridging the Semantic Gap in Categorical Clustering}
\author{Zihua Yang\inst{1} \and
Xin Liao\inst{1} \and
Yiqun Zhang\inst{1,2}\thanks{Corresponding author} \and
Yiu-ming Cheung\inst{2*}}
\authorrunning{Z. Yang et al.}
\institute{School of Computer Science and Technology, \\Guangdong University of Technology, Guangzhou, China \and
Department of Computer Science,\\ Hong Kong Baptist University, Hong Kong SAR, China\\
\email{3122004153@mail2.gdut.edu.cn, 3123004616@mail2.gdut.edu.cn,}\\
\email{yqzhang@gdut.edu.cn, ymc@comp.hkbu.edu.hk}}
\maketitle              
\begin{abstract}

Qualitative data are widespread in domains such as healthcare, marketing, and bioinformatics, where clustering offers a fundamental tool for pattern discovery. A core difficulty of qualitative-data clustering lies in measuring similarity among attribute values that carry no inherent ordering or distance. To recover such relationships, existing studies typically rely on within-dataset co-occurrence statistics. This statistical route, however, becomes unreliable once the sample size is small, and the semantic context of each value is therefore left underexploited.
Motivated by this limitation, this paper proposes BREVE (\textbf{B}alanced \textbf{R}epresentation via \textbf{E}xternal \textbf{V}alue \textbf{E}nrichment), a clustering framework that enriches each qualitative value with extra semantic dimensions drawn from an external knowledge base. That is, every unique value is expanded by a dense embedding that encodes its semantic content. To prevent the original value identity from being diluted by the added dimensions, a lightweight one-hot component is further appended. An adaptive weight, guided by cluster compactness, then determines how strongly the enrichment dimensions enter the final representation. With this design, experiments on eight benchmark datasets yield an average ARI rank of 1.3 against seven representative competitors.
\end{abstract}

%


\section{Introduction}

Qualitative data are pervasive in real-world applications such as medical diagnosis, customer segmentation, and biological research~\cite{chen2024qgrl}. Within these scenarios, discovering latent patterns through clustering supports a series of downstream tasks, e.g., patient stratification, market analysis, and gene function annotation. As an unsupervised paradigm, clustering recovers hidden structures directly from raw observations and avoids the high cost of manual labeling. Unlike numerical data, which inherit a natural metric from their underlying continuous space, qualitative data exist as discrete symbols and lack any built-in notion of distance. Therefore, how to expose the relationships among attribute values has long been the central question in qualitative-data clustering.

To bridge this gap, mainstream studies extract value relationships from statistical regularities within the dataset itself. Distance-based methods, e.g., $k$-modes~\cite{huang1998kmodes}, treat each attribute mismatch uniformly, whereas more recent designs~\cite{feng2025robust} model attribute coupling or fuse multiple metric spaces to capture finer dependencies. Embedding-based methods take an alternative path and learn continuous vectors that reflect value co-occurrence. Different as these designs are, they share a common premise, i.e., the information required to relate values can be fully recovered from internal statistics. The premise, however, no longer holds once the sample size is limited. In specialized scenarios such as rare-disease diagnosis or niche-market analysis, a dataset often holds only tens to hundreds of records, in which case co-occurrence signals are too sparse to separate semantically related values. Consequently, the representation learned from purely internal statistics tends to be under-informed, and the resulting partitions degrade accordingly.

A more promising direction is therefore to introduce external evidence as a supplement. In fact, qualitative attribute values are typically short text tokens whose meanings have already been documented in widely available knowledge bases. Such bases offer a low-cost source of value-level descriptions. Building on this resource, a dense embedding can be retrieved for every unique value and appended to the original representation as additional dimensions, so that semantic affinity left undiscovered by the small dataset is supplied externally. Notably, the discrete identity of each value remains informative for partitioning and should not be diluted by the added dimensions. A balancing mechanism is thus required, so that the enrichment dimensions and the original identity contribute to clustering in a controlled proportion.

Motivated by the above analysis, this paper proposes BREVE (\textbf{B}alanced \textbf{R}epresentation via \textbf{E}xternal \textbf{V}alue \textbf{E}nrichment), a clustering framework that performs value-level dimension enrichment for qualitative data. Specifically, every unique value is enriched with a dense semantic embedding retrieved from an external knowledge base, and a one-hot component is appended to retain its original identity. An adaptive weight, guided by cluster compactness, then regulates how strongly the enrichment dimensions enter the final representation, and a partitional clustering is conducted on the resulting representation. The main contributions are summarized as follows.

\begin{enumerate}
    \item A value-level dimension-enrichment framework is proposed for qualitative-data clustering, which retrieves external semantic descriptions for each unique value and converts them into additional representation dimensions, thereby supplementing the limited statistical signals available in small datasets.
    
    \item A balancing scheme is designed to combine the enrichment dimensions with a lightweight one-hot component that preserves the original value identity. An adaptive weight, driven by cluster compactness, further controls the proportion contributed by the enrichment part, so that the final representation reflects semantic affinity without losing discrete distinctions among values. 
    
    \item Extensive experiments on eight benchmark datasets show that the proposed framework attains an average ARI rank of 1.3 against seven representative competitors, confirming the benefit of value-level enrichment with external knowledge for qualitative-data clustering.
\end{enumerate}

\section{Related Work}
\subsection{Qualitative Data Representation and Clustering}

The discrete nature of qualitative attributes, together with the absence of any built-in order or distance, calls for an explicitly constructed similarity measure. Since the only signal directly available is the data itself, existing studies rely on statistical regularities observed within the dataset to build such a measure, and the resulting designs split into two complementary branches. The first branch focuses on distance construction. Early efforts, e.g., $k$-modes~\cite{huang1998kmodes}, treat each attribute mismatch in a uniform manner, after which information-theoretic formulations replace plain matching to capture finer dissimilarity. Building on this trend, graph-based modeling~\cite{zhang2023graph} encodes value co-occurrence as structural relations, learnable intra-attribute weighting~\cite{zhang2022learnable} adjusts the importance of individual values, and multi-metric fusion~\cite{feng2025robust} blends several metric spaces for a more robust dissimilarity. The second branch turns to representation learning and maps discrete symbols into continuous vectors. For heterogeneous and mixed-type cases in particular, Het2Hom~\cite{zhang2022het2hom} and QGRL~\cite{chen2024qgrl} project attributes into a homogeneous concept space or a quaternion graph, whereas adaptive partition strategies~\cite{zhang2025adaptive} cluster discrete records through hierarchical merging.

Despite the steady progress on the methodology side, all the above efforts share a common assumption, i.e., the relationships among values can be fully recovered from regularities inside the data. As pointed out by a recent survey~\cite{dinh2024survey}, similarity construction for qualitative data is still considered an open problem in the absence of any intrinsic metric. The difficulty is amplified once only a handful of records are available, where co-occurrence signals fail to support a stable inference of value relationships.

\subsection{External Resources for Clustering}

Beyond inferring value relationships from the dataset, another line seeks supplementary information from sources external to the target data. For text clustering, recent works~\cite{zhang2023clusterllm,viswanathan2024fewshot} draw on the semantic priors carried by pre-trained models to refine cluster boundaries or to generate side constraints that better reflect inter-instance affinity. For tabular scenarios, a serialization paradigm~\cite{hegselmann2023tabllm,borisov2023great} converts each record into a natural-language sentence so that such priors can be exploited, whereas joint pre-training frameworks~\cite{yin2020tabert,herzig2020tapas} encode textual and tabular content together for tasks that demand schema-context alignment.

The above attempts confirm the practical value of external priors, yet none of them transfers cleanly to unsupervised qualitative-data clustering. In particular, the text-clustering line operates on natural-language documents instead of short discrete values. The tabular line, on the other hand, is largely supervised and depends on row-level serialization whose cost grows with the sample size, which is undesirable for iterative clustering. In addition, once external descriptions are folded into a representation, common pooling choices, e.g., mean pooling~\cite{reimers2019sentence} or CLS tokens~\cite{devlin2019bert}, treat all positions equally and easily blur the tokens that actually carry discriminative information. How to bring external semantics into qualitative-data clustering in a manner that suits its discrete and small-sample nature, therefore, remains an open problem.

\section{Proposed Method}

Given a qualitative dataset $\mathcal{X} = \{x_1, \ldots, x_N\}$ in which every object is described by $M$ attributes drawn from the value set $\mathcal{V} = \bigcup_{j=1}^{M} V_j$, the clustering task amounts to learning a mapping $\Phi: \mathcal{X} \rightarrow \mathbb{R}^{D}$ whose induced distances reflect the similarity among objects. The discrete nature of $\mathcal{V}$, however, leaves the values without any natural geometric arrangement, so the representation has to be constructed from the symbolic data themselves. To turn each raw symbol into a clustering-friendly vector, three coupled questions arise, i.e., where the semantic content unavailable from co-occurrence may come from, how the resulting textual descriptions of variable length can be compressed into compact and discriminative embeddings, and how these embeddings can be combined with the original value information without one side overwhelming the other. The three questions form a chain in which every later step depends on the output of the previous one, since the quality of retrieved content shapes the encoded embedding, and the embedding in turn decides whether the subsequent balancing operates on a meaningful basis. Notation used throughout the paper is collected in Table~\ref{tab:symbols}.

\begin{table}[t]
\centering
\caption{Summary of key notations.}
\label{tab:symbols}
\begin{tabular}{l|l}
\toprule
Symbol & Description \\
\midrule
$\mathcal{X}$, $N$, $M$ & Qualitative dataset, number of objects, number of attributes \\
$V_j$, $\mathcal{V} = \bigcup_{j} V_j$ & Value domain of $A_j$ and the overall vocabulary \\
$\mathcal{K}$, $\mathcal{E}$ & External knowledge base, pre-trained text encoder \\
$\mathcal{P}$, $T_v$ & Structured prompt and the description returned for value $v$ \\
$e_v \in \mathbb{R}^{d}$, $a_t$ & Embedding of value $v$, attention weight on the $t$-th token \\
$E^{enr} \in \mathbb{R}^{N \times Md}$ & Enrichment representation built from $\{e_v\}$ \\
$E^{id} \in \mathbb{R}^{N \times d_s}$ & One-hot identity matrix, $d_s = |\mathcal{V}|$ \\
$\alpha \in [0,1]$, $Z \in \mathbb{R}^{N \times D}$ & Balancing weight, fused representation, $D = d_s + Md$ \\
$K$, $\mathcal{Y}$ & Number of clusters, cluster assignment \\
\bottomrule
\end{tabular}
\end{table}


\subsection{Enriching Each Value with External Descriptions}

A qualitative attribute often carries an ordinal or contextual structure that frequency counts within the dataset cannot reveal, e.g., the implicit ordering between \emph{low} and \emph{high}. Such latent structure has to come from somewhere outside the data itself, and a natural source is an external knowledge base $\mathcal{K}$ that already encodes how common values relate to each other. Querying $\mathcal{K}$ for a short textual description of each qualitative value provides the raw material from which extra semantic dimensions are later built.

The next decision is whether to issue one query per record or one per unique value. Continuous attributes typically take a different value at every record, whereas qualitative attributes show heavy redundancy, with the same value recurring in many records of the dataset. Querying record by record would therefore inflate the cost without adding new semantic content, and would also risk returning slightly different descriptions for the very same value. Retrieval is consequently performed on the unique vocabulary $\mathcal{V}$, with a structured prompt $\mathcal{P}$ that takes the value $v$, its host attribute $A_j$ and the domain $V_j$ as inputs, so that the description for value $v$ is obtained as
\begin{equation}
\label{eq:retrieval}
    T_v = \mathcal{K}(\mathcal{P}(v, A_j, V_j)),
\end{equation}
yielding the description set $\mathcal{T} = \{T_v : v \in \mathcal{V}\}$ that feeds the encoding stage.

This value-level choice also keeps the retrieval cost manageable. Let $\mathcal{C}_{q}$ denote the unit cost of one query. Value-level retrieval incurs $\mathcal{O}(|\mathcal{V}| \cdot \mathcal{C}_{q})$ in total, in contrast to $\mathcal{O}(N \cdot M \cdot \mathcal{C}_{q})$ required by record-level retrieval, and the relative saving therefore takes the form
\begin{equation}
    \rho \;=\; 1 - \frac{|\mathcal{V}|}{N \cdot M},
\end{equation}
which approaches unity on standard qualitative benchmarks where $N \cdot M \gg |\mathcal{V}|$. Consequently, the external retrieval contributes only a marginal overhead to the overall pipeline.

A final issue concerns the wording of the descriptions. Without explicit guidance, the returned text tends to interleave the discriminative cue of a value with surrounding generic content, which dilutes the signal that the encoder eventually consumes. The prompt $\mathcal{P}$ is therefore organized into a few short slots targeting distinguishing aspects of the value, so that each produced description remains compact and stays focused on what separates one value from another within the same attribute.

\subsection{Compressing Descriptions into Discriminative Embeddings}

The descriptions retrieved in the previous stage are textual sequences of variable length, whereas a clustering algorithm requires every value to be encoded as a fixed-length vector. A pre-trained text encoder $\mathcal{E}$ is therefore applied to map each description $T_v$ into a sequence of token representations
\begin{equation}
    [h_1, \ldots, h_L] = \mathcal{E}(T_v) \in \mathbb{R}^{L \times d},
\end{equation}
after which a single $d$-dimensional embedding $e_v$ has to be derived from these $L$ tokens.

A common practice for such derivation is to take the CLS-token output or to average all token representations uniformly. Neither option, however, fits the present setting well. CLS-based aggregation packs the entire semantic content into one designated slot and tends to underweight locally informative tokens, whereas uniform averaging assigns equal weight to genuinely distinguishing words and to non-informative ones, which dilutes the signal that separates one value from another. A weighting that varies with token importance is therefore needed.

Recent studies on Transformer interpretability~\cite{kobayashi2020attention,modarressi2022globenc} pointed out that attention weights alone do not fully capture how much a token contributes to a sequence representation, and the magnitude of its hidden state often serves as an equally informative signal, since semantically prominent tokens typically respond more strongly across hidden dimensions. Following this, the mean activation of the $t$-th token is taken as a parameter-free indicator of its importance:
\begin{equation}
\label{eq:activation}
    s_t = \frac{1}{d} \sum_{k=1}^{d} h_{t,k},
\end{equation}
and the value-level embedding is then formed as a softmax-weighted combination of all tokens:
\begin{equation}
\label{eq:embedding}
    a_t = \frac{\exp(s_t)}{\sum_{l=1}^{L} \exp(s_l)}, \qquad e_v = \sum_{t=1}^{L} a_t \cdot h_t \in \mathbb{R}^{d}.
\end{equation}
The weighting reduces to plain mean pooling when the scores $\{s_t\}$ are roughly uniform, and concentrates on the most informative tokens when a few of them stand out. Notably, the whole mechanism introduces no learnable parameter, which is desirable on small qualitative datasets where any extra trainable component would invite overfitting.

With one embedding obtained for every value in $\mathcal{V}$, the representation of an object $x_i = [x_{i,1}, \ldots, x_{i,M}]$ is constructed by concatenating the embeddings of its attribute values in a fixed attribute order,
\begin{equation}
    E_i^{enr} = e_{x_{i,1}} \oplus \cdots \oplus e_{x_{i,M}} \in \mathbb{R}^{Md},
\end{equation}
where attribute $A_j$ occupies the segment $[(j-1)d+1, \, jd]$, so that the dimensions belonging to different attributes remain separated and no cross-attribute interference is introduced. Stacking $E_i^{enr}$ over all $N$ objects yields:
\begin{equation}
    E^{enr} = [E_1^{enr}; \, \ldots; \, E_N^{enr}] \in \mathbb{R}^{N \times Md},
\end{equation}
which expands the original $M$ qualitative attributes into a semantic space.

\begin{algorithm}[t]
\caption{The BREVE Framework}
\label{alg:breve}
\begin{algorithmic}[1]
\REQUIRE Dataset $\mathcal{X}$, attributes $\{A_j\}_{j=1}^{M}$, number of clusters $K$, knowledge base $\mathcal{K}$, text encoder $\mathcal{E}$, candidate grid $\mathcal{G}$
\ENSURE Cluster assignments $\mathcal{Y}$

\STATE \textsc{// Retrieval of value-level descriptions}
\FOR{each $v \in \mathcal{V}$ with host attribute $A_j$ and domain $V_j$}
    \STATE Obtain description $T_v \leftarrow \mathcal{K}(\mathcal{P}(v, A_j, V_j))$ by Eq.~\eqref{eq:retrieval}
\ENDFOR

\STATE \textsc{// Encoding into value-level embeddings}
\FOR{each $v \in \mathcal{V}$}
    \STATE Compute token activations $\{s_t\}$ and weights $\{a_t\}$ from $\mathcal{E}(T_v)$, then form $e_v$ by Eqs.~\eqref{eq:activation}--\eqref{eq:embedding}
\ENDFOR
\FOR{$i = 1$ \TO $N$}
    \STATE Build $E_i^{enr} \leftarrow e_{x_{i,1}} \oplus \cdots \oplus e_{x_{i,M}}$ and $E_i^{id} \leftarrow \mathbf{1}_{x_{i,1}} \oplus \cdots \oplus \mathbf{1}_{x_{i,M}}$
\ENDFOR

\STATE \textsc{// Balancing and clustering}
\STATE Apply column-wise z-score normalization to obtain $\hat{E}^{id}$ and $\hat{E}^{enr}$
\STATE Initialize $\alpha^{\star} \leftarrow 0$, \, $S^{\star} \leftarrow -\infty$
\FOR{each $\alpha \in \mathcal{G}$}
    \STATE Form fused representation $Z_\alpha \leftarrow (1-\alpha)\hat{E}^{id} \oplus \alpha \hat{E}^{enr}$ by Eq.~\eqref{eq:fusion}
    \STATE Run $\mathcal{Y}_\alpha \leftarrow k\text{-Means}(Z_\alpha, K)$ and evaluate $S(Z_\alpha, \mathcal{Y}_\alpha)$
    \IF{$S(Z_\alpha, \mathcal{Y}_\alpha) > S^{\star}$}
        \STATE Update $\alpha^{\star} \leftarrow \alpha$, \, $S^{\star} \leftarrow S(Z_\alpha, \mathcal{Y}_\alpha)$
    \ENDIF
\ENDFOR
\RETURN $\mathcal{Y} \leftarrow k\text{-Means}(Z_{\alpha^{\star}}, K)$
\end{algorithmic}
\end{algorithm}

\subsection{Balancing Original Identity and Enriched Dimensions}

The enrichment representation $E^{enr}$ describes each value through its retrieved semantics and brings together values that share similar meanings. Such grouping, however, can also pull originally distinct values closer than they should be, especially when the external descriptions over-generalize and assign near-identical content to surface forms that the dataset still treats as separate categories. Keeping the discrete identity of each value visible after the enrichment is therefore important, so that values left apart by the data are not implicitly merged in the representation.

A direct way of preserving such identity would be to learn a low-dimensional embedding for every value, yet introducing trainable parameters fits poorly with the small-sample setting that motivates the present design. One-hot encoding offers a parameter-free alternative whose orthogonality guarantees that any two distinct values stay linearly separable. Concretely, for an object $x_i$, the one-hot blocks of its $M$ attribute values are concatenated into
\begin{equation}
    E_i^{id} = \mathbf{1}_{x_{i,1}} \oplus \cdots \oplus \mathbf{1}_{x_{i,M}} \in \mathbb{R}^{d_s}, \qquad d_s = |\mathcal{V}|,
\end{equation}
where $\mathbf{1}_v \in \mathbb{R}^{|\mathcal{V}|}$ is the one-hot vector of value $v$. Stacking $E_i^{id}$ across the $N$ objects gives the identity matrix $E^{id} \in \mathbb{R}^{N \times d_s}$, which keeps the original value distinctions intact regardless of any semantic affinity that the enrichment may have introduced.

The next question is how the identity dimensions and the enriched ones should be combined. Adding them element-wise would force both into the same coordinate axes and blur the boundary between the two sources, whereas concatenating them along the feature dimension keeps each source addressable on its own block of coordinates. After column-wise z-score normalization on $E^{id}$ and $E^{enr}$, the fused representation is accordingly obtained as
\begin{equation}
\label{eq:fusion}
    Z_\alpha = (1-\alpha) \cdot \hat{E}^{id} \;\oplus\; \alpha \cdot \hat{E}^{enr} \;\in\; \mathbb{R}^{N \times D}, \qquad D = d_s + Md,
\end{equation}
where $\alpha \in [0,1]$ controls how much weight the enriched dimensions carry relative to the original identity. Intuitively, a small $\alpha$ trusts the original value distinctions, whereas a large $\alpha$ relies more on the externally retrieved semantics.

In the absence of labels, an internal cluster-quality criterion serves as a proxy for selecting $\alpha$. The Silhouette Score
\begin{equation}
    S(Z_\alpha, \mathcal{Y}_\alpha) = \frac{1}{N} \sum_{i=1}^{N} \frac{b_i - a_i}{\max(a_i, b_i)}
\end{equation}
quantifies, for each object, the gap between its mean distance to its own cluster ($a_i$) and the mean distance to its nearest competing cluster ($b_i$), with a higher value indicating a tighter and better-separated partition. Sweeping $\alpha$ over a small candidate grid $\mathcal{G}$, the weight that yields the best partition is taken as
\begin{equation}
    \alpha^* = \arg\max_{\alpha \in \mathcal{G}} \, S(Z_\alpha, \mathcal{Y}_\alpha).
\end{equation}
The selected $\alpha^*$ also carries a transparent interpretation. A value close to $0$ suggests that the original identity already organizes the data well, whereas a value close to $1$ indicates that the enriched dimensions contribute the dominant part of the discrimination.

The clustering is finally performed on $Z_{\alpha^*}$ via $k$-Means, which minimizes
\begin{equation}
    \mathcal{L} = \sum_{k=1}^{K} \sum_{x_i \in C_k} \|\Phi(x_i) - \mu_k\|^2, \quad \text{s.t.} \quad \mu_k = \frac{1}{|C_k|} \sum_{x_i \in C_k} \Phi(x_i),
\end{equation}
where $\Phi(x_i) = (Z_{\alpha^*})_i$ denotes the $i$-th row of the fused representation. The complete procedure is summarized in Algorithm~\ref{alg:breve}.

\begin{theorem}
\label{thm:time}
\textbf{Time Complexity.} The time complexity of BREVE is $\mathcal{O}(|\mathcal{V}| \cdot \mathcal{C}_{q} + N M d + |\mathcal{G}| \cdot N K D)$, where $\mathcal{C}_{q}$ denotes the unit cost of one query to the external knowledge base, $d$ is the hidden dimension of the text encoder, and $D = d_s + Md$ is the dimensionality of the fused representation.
\end{theorem}

\begin{proof}
The pipeline consists of three stages whose costs can be analyzed in turn. In the retrieval stage, the description $T_v$ is produced once per unique value, so this stage runs in $\mathcal{O}(|\mathcal{V}| \cdot \mathcal{C}_{q})$. In the encoding stage, every value embedding $e_v$ is computed once and then reused, after which assembling the object-level vector $E_i^{enr}$ amounts to concatenating $M$ pre-computed $d$-dimensional blocks for each of the $N$ objects, so the cost of this stage is $\mathcal{O}(N M d)$. In the fusion stage, the Silhouette Score is evaluated for every candidate $\alpha \in \mathcal{G}$, where each evaluation runs $k$-Means on $N$ objects in the $D$-dimensional fused space with $K$ clusters at $\mathcal{O}(N K D)$ per iteration, which gives $\mathcal{O}(|\mathcal{G}| \cdot N K D)$ in total. Summing the three stages together yields the stated bound.
\end{proof}

\begin{theorem}
\label{thm:space}
\textbf{Space Complexity.} The space complexity of BREVE is $\mathcal{O}(|\mathcal{V}| \cdot d + N (d_s + M d))$, with the first term accounting for the cached value-level descriptions and embeddings and the second for the object-level identity, enrichment, and fused matrices.
\end{theorem}

\begin{proof}
The retrieved descriptions and the value-level embeddings $\{e_v\}_{v \in \mathcal{V}}$ are stored once and reused across all $N$ objects, contributing $\mathcal{O}(|\mathcal{V}| \cdot d)$, since the storage scales with the vocabulary instead of the dataset size. On the object side, the identity matrix $E^{id} \in \mathbb{R}^{N \times d_s}$ takes $\mathcal{O}(N d_s)$ and the enrichment matrix $E^{enr} \in \mathbb{R}^{N \times M d}$ takes $\mathcal{O}(N M d)$, while their column-wise concatenation $Z_\alpha \in \mathbb{R}^{N \times D}$ with $D = d_s + M d$ occupies $\mathcal{O}(N D)$ of the same order. Combining the per-vocabulary and per-object contributions yields the stated bound.
\end{proof}

\section{Experiments}
\label{sec:experiments}

The empirical study reported in this section examines BREVE in several aspects. The clustering performance is benchmarked against a set of competing methods, the contribution of each component is then isolated through an ablation study, the runtime behaviour is observed when the input scales along different dimensions. Since recent large language models cover a broad range of everyday concepts that match the kind of attribute values appearing in qualitative datasets, they also serve as a natural choice for instantiating the external knowledge base $\mathcal{K}$ used in BREVE.

\begin{table}[t]
\centering
\caption{\textbf{Statistics of the evaluated datasets.} $N$: number of instances; $M$: number of attributes; $K$: number of classes; $|\mathcal{V}|$: number of unique values across all attributes.}
\label{tab:datasets}
\setlength{\tabcolsep}{8pt}
\begin{tabular}{cl c rrrr}
\toprule
\textbf{No.} & \textbf{Dataset} & \textbf{Abbr.} & $\boldsymbol{N}$ & $\boldsymbol{M}$ & $\boldsymbol{K}$ & $\boldsymbol{|\mathcal{V}|}$ \\
\midrule
1 & Soybean          & SB &     307 & 35 & 19 & 133 \\
2 & Mushroom         & MU &   8,124 & 22 &  2 & 126 \\
3 & Breast Cancer    & BC &     286 &  9 &  2 &  51 \\
4 & Zoo              & ZO &     101 & 16 &  7 &  36 \\
5 & Solar Flare      & SF &   1,066 & 10 &  6 &  31 \\
6 & Dermatology      & DE &     366 & 34 &  6 & 133 \\
7 & Lymphography     & LY &     148 & 18 &  4 &  59 \\
8 & Car Evaluation   & CA &   1,728 &  6 &  4 &  21 \\
\bottomrule
\end{tabular}
\end{table}

\subsection{Experimental Settings}

\textbf{Datasets.} Eight qualitative benchmarks from the UCI Machine Learning Repository\footnote{\url{https://archive.ics.uci.edu/}}~\cite{dua2019uci} are adopted, with their basic statistics summarized in Table~\ref{tab:datasets}. The selection spans dataset sizes from $101$ to $8{,}124$ instances and covers a range of application domains, e.g., biological taxonomy, medical diagnosis, and consumer evaluation. Notably, five out of the eight datasets contain fewer than $500$ instances, which echoes the practical situation that qualitative data often arises in specialized scenarios with limited sample availability and therefore offers a fair testbed for the proposed framework.

\textbf{Compared methods.} Two classical baselines and five recent competitors are included for comparison. The classical references are KMD ($k$-Modes)~\cite{huang1998kmodes} and OHK (One-Hot $k$-Means), both of which treat attribute values as plain symbols without modeling their relations. The five recent competitors cover a diverse spectrum of design philosophies, including order-constrained structure learning by COForest~\cite{zhao2024learning}, multi-granularity competitive partitioning by MCDC~\cite{cai2024robust}, significance-based recursive splitting by SigDT~\cite{hu2025significance}, conditional-entropy-driven cluster-specific distance learning by DiSC~\cite{zhao2025break}, and value-order optimization for ordinal-aware distance computation by OCL~\cite{zhang2025categorical}.

\textbf{Evaluation metrics.} Three standard clustering measures are reported. Adjusted Rand Index (ARI)~\cite{gates2017impact} measures the chance-corrected pairwise agreement between the obtained partition and the ground truth. Normalized Mutual Information (NMI)~\cite{estevez2009normalized} quantifies the information shared between the cluster assignment and the class labels. Clustering Accuracy (ACC)~\cite{he2005laplacian} reports the best-match classification accuracy under the optimal label permutation.

\textbf{Implementation details.} Every experiment is repeated $10$ times with different random initializations, and the mean and standard deviation are reported. Statistical significance against the strongest competitor is assessed via the Wilcoxon signed-rank test at the $p<0.05$ level. The text encoder $\mathcal{E}$ is instantiated as all-mpnet-base-v2 throughout. The external knowledge base $\mathcal{K}$ is realized through publicly available large language models, and four mainstream options, i.e., GPT-5.1, Claude Opus 4.5, DeepSeek V3.2, and Gemini 3 Pro, are evaluated to verify that BREVE is not tied to any specific provider.

\subsection{Clustering Performance Evaluation}
The quantitative results on the eight benchmarks are summarized in Table~\ref{tab:main_results}. Across the three measures, BREVE secures the best or second-best score in $21$ out of $24$ dataset-metric combinations. The advantage over the strongest competitor, SigDT, is statistically significant at $p<0.05$ on ARI and NMI according to the Wilcoxon signed-rank test.

\begin{table*}[!t]
\centering
\caption{\textbf{Clustering results on eight qualitative benchmarks.} 
``$^\ast$'' marks statistical significance ($p<0.05$) under the Wilcoxon signed-rank test. 
}
\label{tab:main_results}
\renewcommand{\arraystretch}{1.08}
\setlength{\tabcolsep}{2.5pt}
\resizebox{\textwidth}{!}{%
\begin{tabular}{c|c|ccccccc|cccc}
\toprule
\multirow{2}{*}{\textbf{Metric}} & \multirow{2}{*}{\textbf{Data}}
& MCDC & KMD & COForest & OHK & OCL & DiSC & SigDT
& \multicolumn{4}{c}{\textbf{BREVE (Ours)}} \\
& & {\scriptsize[ICDCS'24]} & {\scriptsize[DMKD'98]} & {\scriptsize[ECAI'24]} & -- & {\scriptsize[SIGMOD'26]} & {\scriptsize[AAAI'26]} & {\scriptsize[Inf.Sci.'25]}
& \textbf{GPT} & $\Delta$\textbf{DS} & $\Delta$\textbf{GE} & $\Delta$\textbf{CL} \\
\midrule
\multirow{8}{*}{\textbf{ARI}}
& CA & 0.001$_{\pm0.00}$ & 0.025$_{\pm0.03}$ & 0.058$_{\pm0.04}$ & 0.033$_{\pm0.05}$ & \underline{0.065}$_{\pm0.05}$ & 0.029$_{\pm0.06}$ & -0.036$_{\pm0.00}$
    & \textbf{0.085}$_{\pm0.08}$ & \textcolor{pos}{+0.12} & \textcolor{neg}{-0.01} & \textcolor{pos}{+0.13} \\
& ZO & 0.602$_{\pm0.00}$ & 0.588$_{\pm0.09}$ & 0.660$_{\pm0.08}$ & 0.596$_{\pm0.12}$ & 0.653$_{\pm0.13}$ & \underline{0.669}$_{\pm0.13}$ & 0.592$_{\pm0.00}$
    & \textbf{0.794}$_{\pm0.08}$ & \textcolor{pos}{+0.01} & \textcolor{neg}{-0.00} & \textcolor{neg}{-0.04} \\
& MU & 0.073$_{\pm0.00}$ & 0.264$_{\pm0.07}$ & 0.376$_{\pm0.22}$ & 0.245$_{\pm0.05}$ & \underline{0.474}$_{\pm0.20}$ & 0.351$_{\pm0.25}$ & 0.339$_{\pm0.00}$
    & \textbf{0.596}$_{\pm0.02}$ & \textcolor{neg}{-0.08} & \textcolor{neg}{-0.03} & \textcolor{neg}{-0.05} \\
& LY & -0.010$_{\pm0.00}$ & 0.028$_{\pm0.04}$ & 0.153$_{\pm0.03}$ & 0.115$_{\pm0.02}$ & 0.131$_{\pm0.05}$ & 0.121$_{\pm0.07}$ & \underline{0.161}$_{\pm0.00}$
    & \textbf{0.204}$_{\pm0.03}$ & \textcolor{pos}{+0.02} & \textcolor{pos}{+0.02} & \textcolor{pos}{+0.03} \\
& DE & 0.641$_{\pm0.00}$ & 0.397$_{\pm0.02}$ & 0.685$_{\pm0.11}$ & 0.487$_{\pm0.21}$ & 0.522$_{\pm0.10}$ & 0.612$_{\pm0.31}$ & \textbf{0.818}$_{\pm0.00}$
    & \underline{0.752}$_{\pm0.07}$ & \textcolor{neg}{-0.01} & \textcolor{neg}{-0.03} & \textcolor{neg}{-0.03} \\
& BC & -0.008$_{\pm0.00}$ & -0.001$_{\pm0.00}$ & 0.029$_{\pm0.06}$ & -0.003$_{\pm0.00}$ & 0.012$_{\pm0.05}$ & 0.004$_{\pm0.03}$ & \underline{0.157}$_{\pm0.00}$
    & \textbf{0.169}$_{\pm0.00}$ & \textcolor{neg}{-0.01} & 0.00 & 0.00 \\
& SF & 0.028$_{\pm0.00}$ & 0.017$_{\pm0.01}$ & 0.009$_{\pm0.03}$ & 0.019$_{\pm0.01}$ & \underline{0.044}$_{\pm0.02}$ & -0.043$_{\pm0.05}$ & \textbf{0.054}$_{\pm0.00}$
    & 0.044$_{\pm0.02}$ & \textcolor{neg}{-0.01} & \textcolor{neg}{-0.01} & \textcolor{neg}{-0.00} \\
& SB & 0.340$_{\pm0.00}$ & 0.322$_{\pm0.04}$ & 0.408$_{\pm0.02}$ & 0.347$_{\pm0.05}$ & 0.387$_{\pm0.04}$ & 0.159$_{\pm0.16}$ & \underline{0.425}$_{\pm0.00}$
    & \textbf{0.425}$_{\pm0.03}$ & \textcolor{pos}{+0.01} & \textcolor{pos}{+0.03} & \textcolor{pos}{+0.03} \\
\midrule
\multicolumn{2}{c|}{Avg.} & 6.9$^\ast$ & 7.4$^\ast$ & 4.6$^\ast$ & 6.5$^\ast$ & 4.1$^\ast$ & 5.9$^\ast$ & 3.4$^\ast$ & \textbf{1.3} & -- & -- & -- \\
\midrule
\multirow{8}{*}{\textbf{NMI}}
& CA & 0.003$_{\pm0.00}$ & 0.047$_{\pm0.02}$ & 0.107$_{\pm0.04}$ & 0.046$_{\pm0.05}$ & \underline{0.126}$_{\pm0.10}$ & 0.052$_{\pm0.06}$ & 0.034$_{\pm0.00}$
    & \textbf{0.153}$_{\pm0.10}$ & \textcolor{pos}{+0.14} & \textcolor{neg}{-0.01} & \textcolor{pos}{+0.16} \\
& ZO & 0.803$_{\pm0.00}$ & 0.763$_{\pm0.03}$ & \underline{0.807}$_{\pm0.04}$ & 0.785$_{\pm0.04}$ & 0.799$_{\pm0.04}$ & 0.794$_{\pm0.06}$ & 0.712$_{\pm0.00}$
    & \textbf{0.859}$_{\pm0.02}$ & \textcolor{neg}{-0.00} & \textcolor{neg}{-0.01} & \textcolor{neg}{-0.01} \\
& MU & 0.059$_{\pm0.00}$ & 0.251$_{\pm0.09}$ & 0.317$_{\pm0.16}$ & 0.230$_{\pm0.03}$ & 0.408$_{\pm0.18}$ & 0.339$_{\pm0.20}$ & \underline{0.436}$_{\pm0.00}$
    & \textbf{0.532}$_{\pm0.03}$ & \textcolor{neg}{-0.05} & \textcolor{neg}{-0.01} & \textcolor{neg}{-0.04} \\
& LY & 0.108$_{\pm0.00}$ & 0.064$_{\pm0.03}$ & 0.181$_{\pm0.03}$ & 0.161$_{\pm0.03}$ & 0.170$_{\pm0.04}$ & 0.174$_{\pm0.06}$ & \underline{0.231}$_{\pm0.00}$
    & \textbf{0.245}$_{\pm0.03}$ & \textcolor{pos}{+0.03} & \textcolor{pos}{+0.01} & \textcolor{pos}{+0.03} \\
& DE & 0.792$_{\pm0.00}$ & 0.634$_{\pm0.01}$ & 0.833$_{\pm0.05}$ & 0.646$_{\pm0.11}$ & 0.712$_{\pm0.05}$ & 0.662$_{\pm0.33}$ & \underline{0.857}$_{\pm0.00}$
    & \textbf{0.867}$_{\pm0.02}$ & \textcolor{pos}{+0.01} & \textcolor{pos}{+0.00} & \textcolor{neg}{-0.00} \\
& BC & 0.001$_{\pm0.00}$ & 0.002$_{\pm0.00}$ & 0.015$_{\pm0.03}$ & 0.004$_{\pm0.00}$ & 0.008$_{\pm0.02}$ & 0.012$_{\pm0.02}$ & \underline{0.070}$_{\pm0.00}$
    & \textbf{0.079}$_{\pm0.00}$ & \textcolor{neg}{-0.00} & 0.00 & 0.00 \\
& SF & \textbf{0.068}$_{\pm0.00}$ & 0.043$_{\pm0.01}$ & 0.046$_{\pm0.01}$ & 0.048$_{\pm0.01}$ & 0.051$_{\pm0.00}$ & 0.033$_{\pm0.01}$ & \underline{0.058}$_{\pm0.00}$
    & 0.050$_{\pm0.01}$ & \textcolor{neg}{-0.00} & \textcolor{neg}{-0.01} & \textcolor{neg}{-0.00} \\
& SB & 0.634$_{\pm0.00}$ & 0.669$_{\pm0.03}$ & 0.707$_{\pm0.02}$ & 0.677$_{\pm0.04}$ & 0.694$_{\pm0.03}$ & 0.329$_{\pm0.33}$ & \underline{0.714}$_{\pm0.00}$
    & \textbf{0.726}$_{\pm0.02}$ & \textcolor{pos}{+0.00} & \textcolor{pos}{+0.01} & \textcolor{pos}{+0.01} \\
\midrule
\multicolumn{2}{c|}{Avg.} & 6.1$^\ast$ & 7.1$^\ast$ & 4.4$^\ast$ & 6.1$^\ast$ & 4.4$^\ast$ & 6.0$^\ast$ & 3.5$^\ast$ & \textbf{1.4} & -- & -- & -- \\
\midrule
\multirow{8}{*}{\textbf{ACC}}
& CA & 0.270$_{\pm0.00}$ & 0.357$_{\pm0.04}$ & 0.402$_{\pm0.05}$ & 0.385$_{\pm0.06}$ & 0.396$_{\pm0.06}$ & \textbf{0.602}$_{\pm0.11}$ & \underline{0.575}$_{\pm0.00}$
    & 0.381$_{\pm0.07}$ & \textcolor{pos}{+0.08} & \textcolor{pos}{+0.01} & \textcolor{pos}{+0.09} \\
& ZO & 0.673$_{\pm0.00}$ & 0.673$_{\pm0.07}$ & 0.695$_{\pm0.07}$ & 0.671$_{\pm0.11}$ & 0.692$_{\pm0.10}$ & 0.710$_{\pm0.09}$ & \underline{0.723}$_{\pm0.00}$
    & \textbf{0.835}$_{\pm0.05}$ & \textcolor{pos}{+0.01} & \textcolor{pos}{+0.01} & \textcolor{neg}{-0.03} \\
& MU & 0.635$_{\pm0.00}$ & 0.475$_{\pm0.00}$ & 0.787$_{\pm0.11}$ & 0.501$_{\pm0.05}$ & \underline{0.830}$_{\pm0.10}$ & 0.773$_{\pm0.11}$ & 0.520$_{\pm0.00}$
    & \textbf{0.886}$_{\pm0.01}$ & \textcolor{neg}{-0.03} & \textcolor{neg}{-0.01} & \textcolor{neg}{-0.02} \\
& LY & 0.446$_{\pm0.00}$ & 0.392$_{\pm0.04}$ & \underline{0.546}$_{\pm0.04}$ & 0.476$_{\pm0.02}$ & 0.484$_{\pm0.06}$ & 0.521$_{\pm0.05}$ & \textbf{0.588}$_{\pm0.00}$
    & 0.540$_{\pm0.02}$ & \textcolor{pos}{+0.03} & \textcolor{pos}{+0.02} & \textcolor{pos}{+0.02} \\
& DE & 0.705$_{\pm0.00}$ & 0.549$_{\pm0.07}$ & 0.739$_{\pm0.09}$ & 0.634$_{\pm0.16}$ & 0.655$_{\pm0.09}$ & 0.702$_{\pm0.20}$ & \textbf{0.833}$_{\pm0.00}$
    & \underline{0.785}$_{\pm0.07}$ & \textcolor{neg}{-0.00} & \textcolor{neg}{-0.02} & \textcolor{neg}{-0.02} \\
& BC & 0.531$_{\pm0.00}$ & 0.533$_{\pm0.02}$ & 0.552$_{\pm0.09}$ & 0.516$_{\pm0.00}$ & 0.529$_{\pm0.07}$ & 0.586$_{\pm0.08}$ & \textbf{0.729}$_{\pm0.00}$
    & \textbf{0.729}$_{\pm0.00}$ & \textcolor{neg}{-0.00} & 0.00 & 0.00 \\
& SF & 0.364$_{\pm0.00}$ & 0.360$_{\pm0.03}$ & 0.373$_{\pm0.02}$ & 0.358$_{\pm0.00}$ & 0.415$_{\pm0.07}$ & \underline{0.452}$_{\pm0.11}$ & 0.356$_{\pm0.00}$
    & \textbf{0.492}$_{\pm0.09}$ & \textcolor{neg}{-0.02} & \textcolor{neg}{-0.01} & \textcolor{pos}{+0.02} \\
& SB & 0.564$_{\pm0.00}$ & 0.530$_{\pm0.00}$ & \underline{0.595}$_{\pm0.04}$ & 0.549$_{\pm0.03}$ & 0.585$_{\pm0.04}$ & 0.323$_{\pm0.17}$ & 0.560$_{\pm0.00}$
    & \textbf{0.656}$_{\pm0.03}$ & \textcolor{pos}{+0.00} & \textcolor{pos}{+0.02} & \textcolor{pos}{+0.02} \\
\midrule
\multicolumn{2}{c|}{Avg.} & 6.0$^\ast$ & 7.1$^\ast$ & 4.0$^\ast$ & 6.6$^\ast$ & 4.9$^\ast$ & 4.3$^\ast$ & 3.0 & \textbf{1.6} & -- & -- & -- \\
\bottomrule
\end{tabular}%
}
\end{table*}

The advantage of BREVE is most pronounced on small-scale datasets such as ZO, LY and BC, where the limited sample size leaves co-occurrence statistics too sparse for reliable value relations. On BC in particular, several competitors including MCDC and OHK fall close to zero in ARI, whereas BREVE substantially raises clustering quality by drawing on the external descriptions retrieved for each value. Such an outcome aligns with the original motivation, since the descriptions returned by $\mathcal{K}$ provide exactly the kind of value-level information that within-dataset statistics fail to expose under tight sample budgets.
Two further benchmarks, CA and SF, mark the situations in which the enrichment strategy reaches its limit. On CA, the highest ARI is obtained by BREVE because the external descriptions surface the ordinal progression among the attribute values. On SF, in contrast, SigDT slightly surpasses BREVE, since the astronomical identifiers used in this dataset carry little public semantic content and an external knowledge base contributes only marginal information beyond what statistical optimization already captures. Overall, this pattern delineates the regime where external semantic enrichment is genuinely beneficial, namely qualitative data whose values correspond to commonly known concepts.

The $\Delta$ columns further report how the choice of back-end for $\mathcal{K}$ affects the obtained partitions. The four candidates produce broadly consistent results, with most $\Delta$ values staying within $\pm0.05$ across datasets, which indicates that BREVE is largely insensitive to the specific provider being queried. A few datasets do show wider gaps, e.g., DeepSeek and Claude improve over GPT by more than $+0.10$ on CA, suggesting that certain providers articulate ordinal concepts in a more clustering-friendly manner. Such stability is jointly supported by the broad concept coverage shared across modern back-ends and by the attention-weighted encoding together with the balancing weight $\alpha$, both of which absorb provider-specific stylistic variation before the clustering stage.

\begin{figure}[t]
    \centering
    \includegraphics[width=\linewidth]{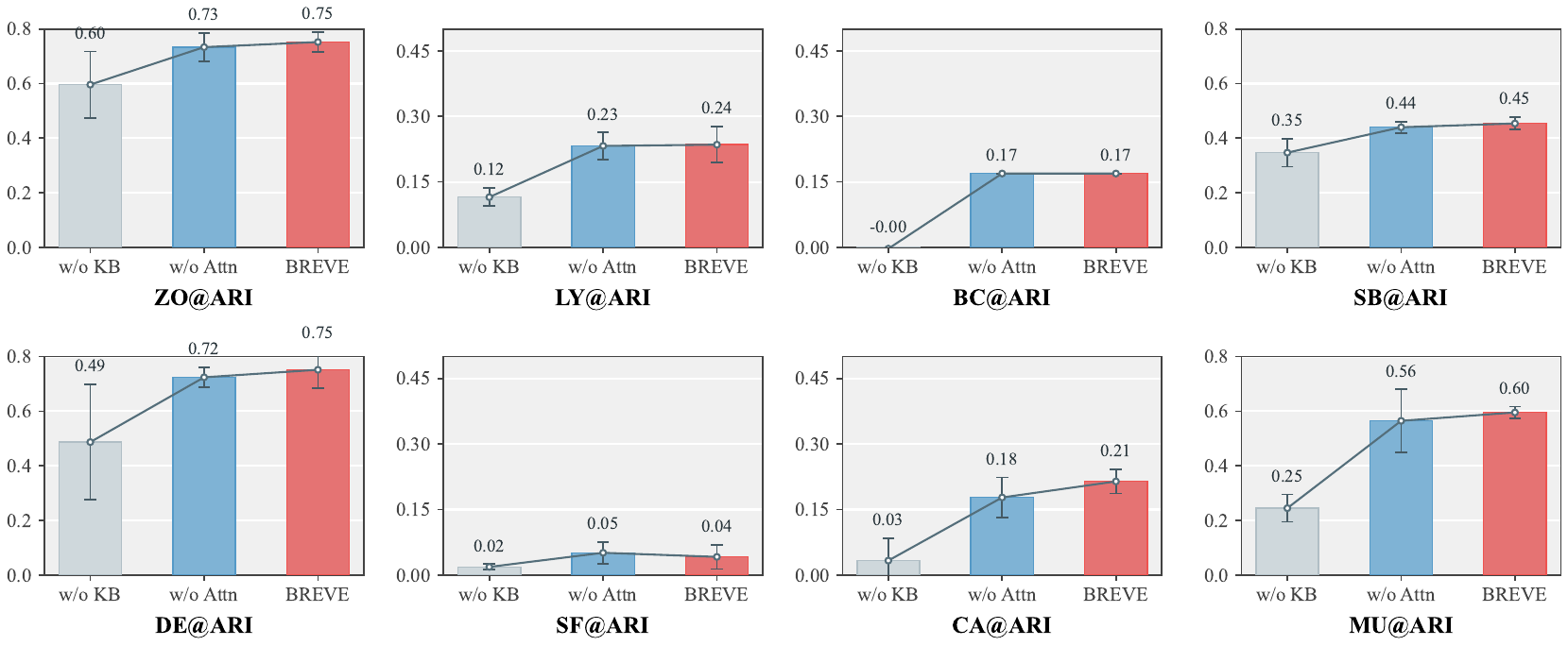}
    \caption{\textbf{Ablation study on the two components of BREVE.} ARI of the identity-only baseline (\textit{w/o KB}), the CLS-pooling variant (\textit{w/o Attn}) and the complete BREVE on the eight benchmarks.}
    \label{fig:ablation}
\end{figure}

\subsection{Ablation Study}


To gauge how much each component of BREVE contributes to the final performance, two reduced variants are compared with the full design on the eight benchmarks. The variant \textit{w/o KB} discards the external knowledge base and runs clustering on the one-hot identity component alone, which isolates the gain attributable to the enrichment dimensions. The variant \textit{w/o Attn} keeps the retrieval but swaps the attention-weighted pooling for the CLS-token alternative, which isolates the contribution of the proposed pooling scheme.

Attaching the retrieved descriptions to the identity component already lifts the average ARI by a sizable margin, as reported in Fig.~\ref{fig:ablation}, and the gain is most pronounced on benchmarks whose values carry strong semantic content, e.g., DE and MU. Such a margin indicates that the enrichment dimensions are the dominant source of BREVE's overall advantage, and that the one-hot component on its own falls noticeably short on every dataset.
Switching the pooling from the CLS token to the attention-weighted scheme then provides an additional but more modest refinement. The improvement is consistent on most benchmarks and is most visible on CA and MU, where the retrieved descriptions contain richer discriminative tokens that benefit from the activation-based reweighting. On BC and SF, by contrast, the two pooling schemes deliver virtually the same ARI, since the values in these benchmarks correspond to specialized medical or astronomical terms whose descriptions show little internal variation. Overall, the two components are clearly complementary, where the external descriptions deliver the dominant share of the improvement and the attention-weighted pooling supplies a final layer of refinement.

\begin{figure}[t]
    \centering
    \includegraphics[width=\linewidth]{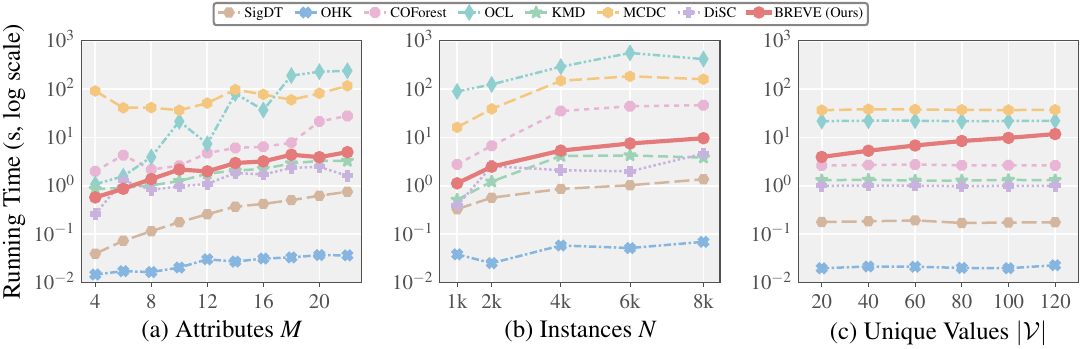}
    \caption{\textbf{Runtime analysis on a log-scale time axis.} Execution time as a function of (a) the attribute count $M$, (b) the instance count $N$, and (c) the vocabulary size $|\mathcal{V}|$. The runtime reported in (c) includes the one-time external description retrieval.}
    \label{fig:scalability}
\end{figure}


\subsection{Scalability Analysis}

Figure~\ref{fig:scalability} shows the runtime with respect to attribute count $M$, instance count $N$, and unique value count $|\mathcal{V}|$.
As shown in Fig.~\ref{fig:scalability}(a), BREVE scales linearly with the number of attributes, since each attribute embedding is retrieved independently. Fig.~\ref{fig:scalability}(b) further indicates that the runtime growth of BREVE with respect to $N$ also remains linear, matching that of classical symbolic methods such as OHK and KMD. Such linear behaviour confirms that the proposed enrichment does not elevate the complexity class. Compared with coupling-based methods such as MCDC, OCL, and COForest, BREVE reduces runtime by orders of magnitude, which results from its decoupled design where the retrieval of external descriptions is isolated to an offline preprocessing phase, avoiding the expensive iterative interactions typical of coupling frameworks.
Fig.~\ref{fig:scalability}(c) reports the total runtime including the offline retrieval, which scales linearly with $|\mathcal{V}|$. The online clustering time of BREVE remains stable across all vocabulary sizes. Notably, since each value is processed independently during the offline stage, the retrieval is naturally parallelizable across providers.



\section{Concluding Remarks}

This paper has presented BREVE, a framework that enriches every unique value of a qualitative dataset with extra semantic dimensions retrieved from an external knowledge base. The variable-length descriptions returned for each value are turned into compact embeddings through an attention-weighted pooling that emphasizes the discriminative tokens. Furthermore, BREVE appends a one-hot identity component to the enriched dimensions and adaptively balances their contributions for clustering, so that the original value information is preserved against the added dimensions. Experiments on eight benchmark datasets across seven competitors validate the effectiveness of BREVE, which attains the best average rank across all the evaluated datasets. Moreover, four mainstream large language models have been adopted to instantiate the knowledge base, with the results staying broadly consistent across providers. The improvements are particularly notable when the data scale is small, which echoes that external value-level enrichment is helpful when within-dataset evidence is scarce. The next avenue of this work could be extending to mixed-type data and tailoring the prompt for domain- and task-specific adaptation.

\subsubsection{\ackname} This work was supported in part by the National Natural Science Foundation of China (NSFC) under grant: 62476063; in part by the NSFC/Research Grants Council (RGC) Joint Research Scheme under grant: N\_HKBU214/21; in part by the General Research Fund of RGC under grants: 12202025 and 12202924; in part by the Natural Science Foundation of Guangdong Province under grant: 2025A1515011293; and in part by the Guangdong and Hong Kong Universities ``1+1+1'' Joint Research Collaboration Scheme with grant: 2025A0505000004.

%
%

\end{document}